\pdfoutput=1

\documentclass[11pt]{article}

\usepackage[]{acl}

\usepackage{times}
\usepackage{latexsym}

\usepackage[T1]{fontenc}

\usepackage[utf8]{inputenc}

\usepackage{microtype}

\usepackage{fdsymbol}
\newcommand\blfootnote[1]{%
  \begingroup
  \renewcommand\thefootnote{}\footnote{#1}%
  \addtocounter{footnote}{-1}%
  \endgroup
}

\usepackage{inconsolata}
\usepackage{graphicx}
\usepackage{multirow}
\usepackage{tabularx}
\usepackage[export]{adjustbox}
\usepackage{caption}
\usepackage{subcaption}
\usepackage{booktabs}
\usepackage{array}
\usepackage{xcolor}
\usepackage{xspace}
\usepackage{xspace}
\usepackage[normalem]{ulem}
\usepackage{algpascal}
\usepackage{inconsolata}
\usepackage{letltxmacro}
\usepackage{booktabs}
\usepackage{algorithm}
\usepackage{algpascal}
\usepackage{algpseudocode}
\usepackage{algorithmicx}
\usepackage{caption}
\usepackage{colortbl,xcolor,xspace}
\usepackage{soul}
\usepackage{nicematrix}
\usepackage{times}
\usepackage{latexsym}
\usepackage{graphicx}
\usepackage{multirow}
\usepackage{tabularx}
\usepackage[export]{adjustbox}
\usepackage{longtable}
\usepackage{xtab,afterpage}
\usepackage[normalem]{ulem}
\title{CoDa: Constrained Generation based Data Augmentation \\
for Low-Resource NLP}

\author{Chandra Kiran Evuru$^{\spadesuit*}$ \quad Sreyan Ghosh$^{\spadesuit*}$ \quad Sonal Kumar$^{\spadesuit}$  \quad Ramaneswaran S$^{\clubsuit}$ \\
\bf  Utkarsh Tyagi$^{\spadesuit}$ \quad
\bf Dinesh Manocha$^{\spadesuit}$  \\
        $^{\spadesuit}$University of Maryland, College Park, USA \quad
         $^{\clubsuit}$NVIDIA, Bangalore, India \\
         \texttt{\{ckevuru, sreyang, sonalkum, utkarsht, dmanocha\}@umd.edu}, 
         \texttt{ramanr@nvidia.com}}

\begin{document}
\maketitle
\begin{abstract}
We present \textbf{CoDa} (\textbf{Co}nstrained Generation based \textbf{Da}ta Augmentation), a controllable, effective, and \textit{training-free} data augmentation technique for low-resource (data-scarce) NLP. Our approach is based on prompting off-the-shelf instruction-following Large Language Models (LLMs) for generating text that satisfies a set of constraints. Precisely, we extract a set of simple constraints from every instance in the low-resource dataset and verbalize them to prompt an LLM to generate novel and diverse training instances. Our findings reveal that synthetic data that follows simple constraints in the downstream dataset act as highly effective augmentations, and CoDa can achieve this without intricate decoding-time constrained generation techniques or fine-tuning with complex algorithms that eventually make the model biased toward the small number of training instances. Additionally, CoDa is the first framework that provides users explicit control over the augmentation generation process, thereby also allowing easy adaptation to several domains. We demonstrate the effectiveness of CoDa across 11 datasets spanning 3 tasks and 3 low-resource settings. CoDa outperforms all our baselines, qualitatively and quantitatively, with improvements of 0.12\%-7.19\%. Code is available~\footnote{https://github.com/Sreyan88/CoDa}.\blfootnote{$^*$These authors contributed equally to this work.}









\end{abstract}






\section{Introduction}
\label{sec:intro}
Data augmentation is a widely used technique to address the problem of limited training data in low-resource NLP~\cite{chen2023empirical}. Owing to the recent advancement in generative AI, using synthetic data to train task-specific models has also gained much popularity. However, given a low-resource NLU dataset, effectively generating task-specific data to expand the dataset still poses a significant challenge. For example, while diversity in tokens and contexts in generated augmentations typically benefits downstream performance, excessively diverse examples may negatively impact consistency with the underlying downstream data distribution, thereby hurting performance~\cite{geiping2023how}. This highlights the importance of having more control during the generation process to ensure data augmentation is done effectively.

\begin{table}[t!]
    \centering
    \resizebox{1.0\columnwidth}{!}{%
    {\renewcommand{\arraystretch}{1.05}%
    \begin{tabular}{ >{\footnotesize} l | >{\small} l }
    
    \toprule
         \multirow{2}{*}{\textbf{Method}}& \textbf{Original 1:} Scott Gimple sets the record straight on that big midseason finale scene.\\
        &  \textbf{Original 2:} mine is all of Isaiah chapter 6 because that is what Jesus did for me.\\
        \midrule
         BackTrans & 1. Scott Gimple is setting the record on this. \\
         \cite{yu2018qanet} & 2: all of Isaiah, chapter 6, is mine, for this is what Jesus. \\
        \midrule
        EDA & 1: scott gimple sets the record straight on that big midseason finale scenery.\\
        \cite{wei2019eda} & 2: mine is all of isaiah chapter because personify that is what jesus did for me.\\
        \midrule
        SSMBA & 1. scott gimple sets the record, on that big preseason night website.  \\
         \cite{ng-etal-2020-ssmba} & 2: it in all of isaiah - 6 because that is what you did for me. \\
        \midrule
        GPT3Mix & 1: Scott Gimple sets the record straight on that big midseason finale scene. \\
         \cite{yoo-etal-2021-gpt3mix-leveraging} & 2: I cherish Isaiah chapter 6 as it embodies what Jesus has done for me. \\
        \midrule
         GENIUS & 1: For the record, there was no midseason finale scene. That is, until the finale.\\
         \cite{guo2022genius} & 2: Isaiah chapter 6 because it says, "If you don't believe in God, you will die."\\
        \midrule
        \multirow{5}{*}{\textbf{CoDa} \emph{\textit{(ours)}}} & 1: The recording of the scene in the new movie by Scott Gimple was a big success,\\ & capturing the essence of the entertaining story.\\
        & 2: Jesus taught us to be free, to follow our hearts and minds, and to live life to the fullest.\\ & In today's society, we must continue to mine the teachings of Jesus to find the courage \\ & to live our lives on our own terms.\\
    \bottomrule
    \end{tabular}%
    }}
    \caption{\small Comparison of augmentations generated using CoDa with our baselines. CoDa generates augmentations that are more coherent and diverse. More examples in Appendix~\ref{sec:qual}.}
    \label{tab:teaser-table}
\end{table}

In the past, researchers have employed methods like text-editing~\cite{wei2019eda,karimi-etal-2021-aeda-easier,shou-etal-2022-amr}, fine-tuning Pre-trained Language Models (PLMs) with various algorithms~\cite{wang-etal-2022-promda,zhou2021melm,guo2022genius,ghosh-etal-2023-dale, ghosh-2023-aclm}, etc. However, most of these methods do not impose explicit controls to achieve diversity or consistency. The recent rise of autoregressive LLMs, known for their advanced generative and reasoning skills, introduces promising yet under-explored opportunities to enhance diversity in task-specific synthetic data synthesis. However, controlling autoregressive generation has proved to be innately challenging and complex~\cite{zhang2023tractable}, and prompting-based methods have often employed manual human efforts for extracting data attributes that promote consistency~\cite{yu2023large}.  


\begin{figure*}[t]
    \centering
    \includegraphics[width=\textwidth]{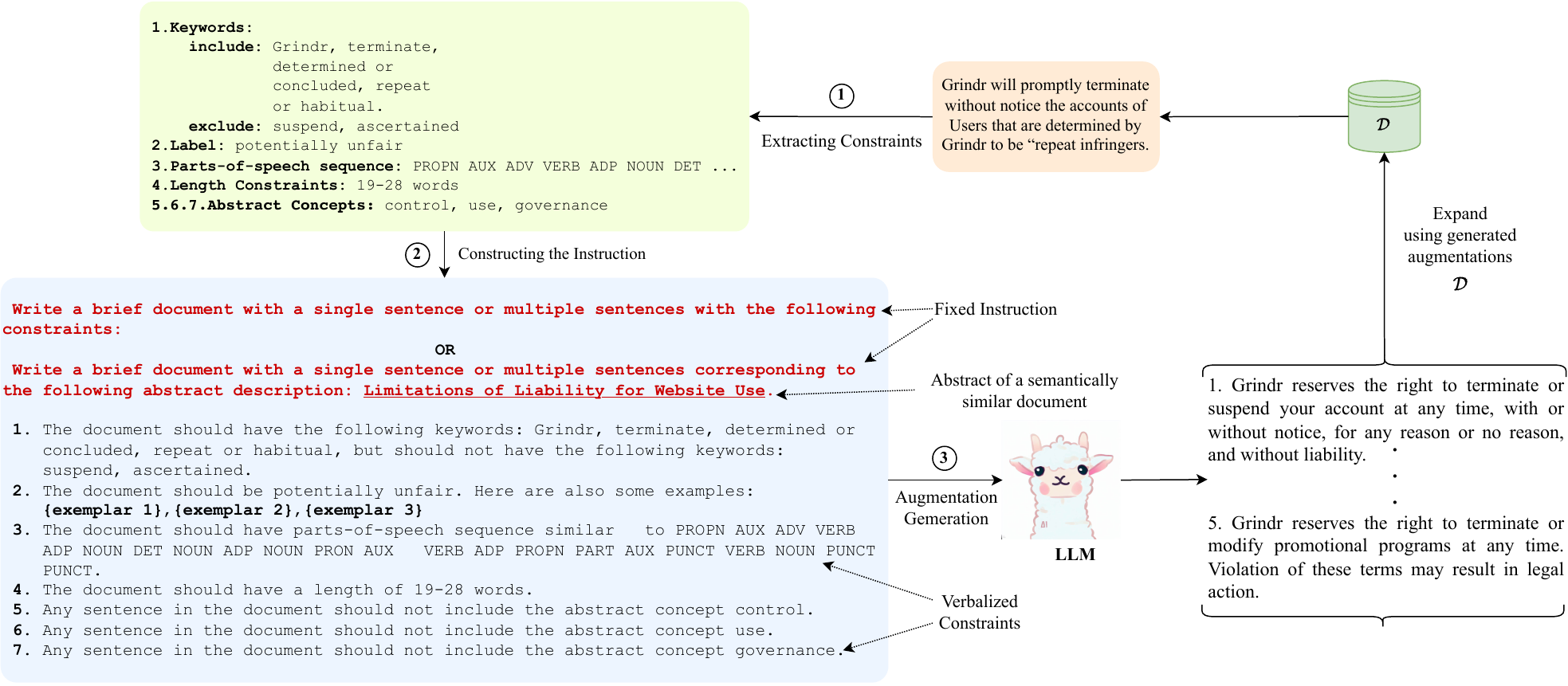}
    \caption{\small Illustration of \textbf{CoDa}. \textcircled{\raisebox{-0.9pt}{1}} For every document in a low-resource NLU dataset $\mathcal{D}$, we extract a set of simple heuristic-based constraints from and \textcircled{\raisebox{-0.9pt}{2}} verbalize them to generate an instruction. \textcircled{\raisebox{-0.9pt}{3}} This instruction is then fed to an existing instruction-tuned LLM for generating augmentations, which are then added to $\mathcal{D}$ for training a downstream model.}
    \label{fig:main_diag}
    \vspace{-1em}
\end{figure*}

\pagebreak

{\noindent \textbf{Main Contributions.}} We propose CoDa, a novel and effective data augmentation methodology for low-resource NLP. CoDa works with any off-the-shelf instruction-tuned LLM in a training-free fashion and provides explicit control over generated augmentations. We first extract simple heuristic-based constraints from training instances in a low-resource NLU dataset and then verbalize them to construct a natural language instruction. Next, we use this instruction to prompt an LLM for generating augmentations (example in Fig.~\ref{fig:main_diag}). Alternative to complex decoding-time-constrained generation methods and manual attribute extraction, CoDa provides a simpler and more intuitive natural language-based interface for constrained generation. CoDa is also the first framework to explore controlled generation for data augmentations, which ensures that the synthetic data is closely aligned with the specific needs of the task and characteristics of the target domain. We show that CoDa, which is \textit{training-free} and much simpler, quantitatively and qualitatively outperforms all prior-art by 0.12\%-7.19\% across various settings.

\section{Related Work}
\label{sec:related_work}

Generative data augmentation for low-resource NLP has been extensively studied in prior work and can be categorized into four primary techniques. Firstly, \textbf{text-infilling} involves corrupting source text segments and using a PLM to refill these gaps. This process often relies on conditioning the corrupted text, a concept also known as keyword conditioning in some studies ~\cite{zhou2021melm,guo2022genius,ghosh-2023-aclm,ghosh-etal-2023-dale,10.1145/3539618.3591957}. Secondly, \textbf{text editing} focuses on modifying certain parts of a given sentence ~\cite{wei2019eda,shou-etal-2022-amr}. Thirdly, \textbf{prompting} involves generating new training sentences by prompting LLMs~\cite{ye-etal-2022-zerogen,sahu2023promptmix}, which can be further uncategorized into conditioning attributes, exemplars, or constraints derived from training data. 



\section{Methodology}
\label{sec:method}
Fig.~\ref{fig:main_diag} illustrates the CoDa pipeline. Given a low-resource dataset $\mathcal{D} = \{d_0, \cdots, d_i, \cdots d_n\}$, we first extract a set of simple heuristic-based constraints from each document $d_i$ and then verbalize the constraints to construct an instruction $I_{d_i}$. After this, we either instruct the LLM with $I_{d_i}$ to generate a completely new document or rephrase another existing document from $\mathcal{D}$. For the latter, we first retrieve a document from $\mathcal{D}$, convert it into its short and concise abstract description by prompting an LLM, and then employ $I_{d_i}$ to generate a document from the abstract description and the extracted constraints. For retrieval, we calculate cosine similarity between SentenceBERT embeddings~\cite{reimers2019sentence} of the source document $d_i$ and all other documents in $\mathcal{D}$, and we randomly sample a sentence from the top-\textit{k} and bottom-\textit{k} similar sentences. For a total of 5 augmentations, we generated 3 novel documents and rephrased 2 other documents for every $d_i$. Finally, all the generated augmentations are added to $\mathcal{D}$ for training a downstream NLU model. We now describe our methodology to extract constraints in detail.

\subsection{Extracting Constraints}
\label{sec:constraints}

{\noindent \textbf{a) Lexical Constraints.}} Inspired by a wealth of prior work in generative data augmentation and constrained generation~\cite{pmlr-v202-zhou23g}, we extract a set of keywords from a source sentence and constrain the augmentations to contain these keywords. More specifically, given a source document $\mathrm{d}$, we first extract all its \textit{n}-grams (1 to 3-grams) $\mathrm{N}$ = \{$\mathrm{n_0}$, $\cdots$, $\mathrm{n_t}$, $\cdots$, $\mathrm{n_T}$\}. Next, we assign an importance score to each by calculating cosine similarity between $\mathbf{E}(\mathrm{n_t})$ and $\mathbf{E}(\mathrm{d})$, where $\mathbf{E}$  is pre-trained SentenceBERT. Finally, we select the top-\textit{k} \textit{n}-grams as our keywords. Additionally, for tasks like NER and QA, we add the corresponding target spans to the list.
\vspace{0.5mm}

{\noindent \textbf{b) Syntactic Constraints.}} In formal domains such as legal and biomedical, language is often governed by syntactical structures. Following a predefined POS pattern ensures that the generated sentences adhere to the formal style and tone expected in the domain. Readers can refer to Appendix~\ref{tab:syntactic_constraints} for some examples. Thus, we consider syntactic constraints that necessitate the generated augmentations to adhere to specific syntactic rules. More specifically, we extract the part-of-speech sequence from a randomly chosen sentence in $\mathrm{d}$ and constrain our generations to adhere to the sequence for a particular sentence.
\vspace{0.5mm}

{\noindent \textbf{c) Semantic (Label) Constraints.}} A primary requirement for effective data augmentations is that the semantics of the generated augmentations adhere to the underlying label of the source document $\mathrm{d}$. To satisfy this, we consider label constraints so that the generated augmentations align closely to the original target label (e.g., \textit{positive} sentiment). We use the target label of $\mathrm{d}$ with 3 exemplars for this constraint. The exemplars are chosen randomly from the dataset $\mathcal{D}$ and placed in random order in the final instruction.
\vspace{0.5mm}

{\noindent \textbf{d) Length  Constraints.}} Length mismatches between training and testing instances have been known to degrade downstream NLU performance~\cite{rogers2021primer}. Motivated by this, we consider length constraints that necessitate the total number of tokens in the generated augmentations to fall within a specified range. We calculate the total number of tokens in $\mathrm{d}$ and add and subtract $sd$ from it to obtain the lower and upper limits of the range, respectively. The value of $sd$ is determined by computing the standard deviation of length distribution across the entire dataset $\mathcal{D}$.

{\noindent \textbf{e) Concept Constraints.}} The presence of spurious features in the training set causes the downstream NLU model to adopt shortcut learning strategies, impacting its performance in real-world, atypical situations where these features are not present~\cite{Sagawa*2020Distributionally}. Data augmentations can further amplify such spurious features in $\mathcal{D}$ if not handled correctly. We propose a novel strategy to ensure that generated augmentations do not have spurious features. We first employ the method proposed by ~\citet{friedman-etal-2022-finding} to extract a list of spurious phrases for each label in the dataset. We then pass these phrases with example sentences consisting of these phrases to an LLM and ask it to return a short abstract concept that the spurious phrases describe in the documents (e.g., \textit{rating in movie reviews} for negative reviews in the IMDB dataset). Finally, we select the top 3 abstract concepts for each label and add is as a negation constraint for augmentation generation.

\vspace{0.5mm}

\begin{table*}[t]
\small
\centering
\setlength{\tabcolsep}{2.8pt}
{\renewcommand{\arraystretch}{1.05}%
\resizebox{\linewidth}{!}{
    \begin{tabular}{l|ccc|ccc|ccc|ccc|ccc}
    \toprule    
    \multicolumn{1}{c|}{\multirow{2}{*}{\textbf{Model}}} &  \multicolumn{3}{c|}{\textbf{Huffpost}}                     & \multicolumn{3}{c|}{\textbf{Yahoo}}                        & \multicolumn{3}{c|}{\textbf{OTS}}                         & \multicolumn{3}{c|}{\textbf{ATIS}} & \multicolumn{3}{c}{\textbf{Massive}} \\  
    \multicolumn{1}{c|}{}                                & \textbf{100} & \textbf{200} & \textbf{500}  & \textbf{100} & \textbf{200} & \textbf{500} & \textbf{100} & \textbf{200} & \textbf{500} & \textbf{100} & \textbf{200} & \textbf{500} & \textbf{100} & \textbf{200} & \textbf{500}  \\ \midrule 
    Gold    & 76.82             & \ul{77.96}             & \ul{80.51}                   & 42.50            & 49.50 &  \ul{55.47}                    & 74.75    & 83.49            & \cellcolor{magenta!20}\textbf{95.14}                        & 85.13             & 89.97             & 94.70 & 31.70  & 56.48 & 73.47        \\
    BackTrans                                            & 75.87             & 76.21             &  79.20                         & 44.85             & 50.86             & 54.19                         & 70.46             & 72.76           & 78.93                        & 89.86             & 92.34             & 94.36   & \ul{53.56}             & 64.52             & 73.13             \\
    EDA                                                  & 75.49             & 77.64             & 79.14                        & 47.13             & 50.15             & 53.39                        & 77.66             & \ul{84.46}            & 87.37                       & 90.20             & 92.11             & 94.93  & 47.00             & 64.15             & 73.53                      \\
    AEDA                                                 &  77.65            & 76.88             & 80.31                        & 45.61             &  51.52            &  54.22                         & 76.56             & 74.75             & 80.92                       &  89.07             & 91.89             & \ul{96.70} & 51.04             & \ul{66.81}              &        \ul{75.15}                        \\
    AMR-DA                                               &77.49              &76.32              &77.93                             &48.80              &\ul{52.37}              &54.68                             &77.98              &78.37              &86.54                                    & \ul{93.69}             & 94.03             & 96.28 & 52.82             & 64.02             & 72.09                  \\
    SSMBA                                                & 76.64             & 77.4             &  79.85                     & 46.95             & 50.53             &  53.97                         & \ul{78.64}             & 83.92            & 85.94                      &  90.31             &  89.75            & 93.69                     &47.07             & 60.99             &        70.24        \\
    GENIUS     & 77.52             & 77.71             & 78.35     & 51.90             & 51.69             & 51.46                         & 77.32             &  75.72           & 78.64                        & 93.58             &  94.14            & \ul{96.70}     & 51.76  & 65.34 & 73.17                 \\
    PromDA     &\ul{77.83}              &77.90              &77.65              &\ul{52.61}               &52.13        &53.40    &78.19   &78.63   &83.69         & 93.49             &  92.76            & 95.11              & 51.68             & 65.71              &74.98     \\
    PromptMix      &-              &-              &-              &-               &-              &-              &-              &-               &-             & 92.68             & \ul{94.25}             & 94.81    & 52.60             & 64.53             & 74.26  \\
    ZeroGen         &73.84              &75.66              &76.30              &41.47               &49.21              &54.55              &68.42              &80.19               &86.79              &81.24              &83.95              &85.63     &28.20              &47.02              &67.80  \\
    GPT3Mix                                             & 57.87            & 61.80             & 66.12 &31.60  & 32.98  & 50.33 & 62.58              & 74.90   & 80.73          & 76.91             & 81.75             &  85.36 & 25.91 & 46.72 & 68.99 \\
    \textbf{CoDa \textit{(ours)}}                                 & \cellcolor{magenta!20}\textbf{79.70}             & \cellcolor{magenta!20}\textbf{80.11}             & \cellcolor{magenta!20}\textbf{81.20}             & \cellcolor{magenta!20}\textbf{53.70}              & \cellcolor{magenta!20}\textbf{54.32}             & \cellcolor{magenta!20}\textbf{55.81}             & \cellcolor{magenta!20}\textbf{84.58}            & \cellcolor{magenta!20}\textbf{86.72}              &\ul{88.63}             & \cellcolor{magenta!20}\textbf{93.92}             & \cellcolor{magenta!20}\textbf{94.45}  & \cellcolor{magenta!20}\textbf{96.82} & \cellcolor{magenta!20}\textbf{54.64} & \cellcolor{magenta!20}\textbf{67.74} & \cellcolor{magenta!20}\textbf{76.20}  \\ 
      & $\pm0.31$           & $\pm0.26$             & $\pm0.11$ & $\pm0.52$ & $\pm0.22$  & $\pm0.31$ & $\pm0.10$              & $\pm0.69$  & $\pm0.45$          & $\pm0.18$             & $\pm0.13$            &  $\pm0.04$ & $\pm0.28$ & $\pm0.15$ & $\pm0.82$\\
    \bottomrule           
    \end{tabular}
}
}
\caption{\small Result comparison for Sequence Classification tasks. CoDa outperforms baselines by 0.12\% - 5.94\%.}
\label{tab:classification}
\end{table*}

\begin{table*}[t]
\begin{minipage}{0.60\textwidth}
\scriptsize
\centering
\resizebox{\textwidth}{!}{
\setlength{\tabcolsep}{1.8pt}
{\renewcommand{\arraystretch}{1.05}%
\begin{tabular}{l|ccc|ccc|ccc|ccc}
\toprule
\multicolumn{1}{c|}{\multirow{2}{*}{\textbf{Model}}} &  \multicolumn{3}{c|}{\textbf{CoNLL-2003}}                     & \multicolumn{3}{c|}{\textbf{OntoNotes}}                        & \multicolumn{3}{c|}{\textbf{EBMNLP}}                         & \multicolumn{3}{c}{\textbf{BC2GM}}\\  
\multicolumn{1}{c|}{} & \textbf{100} & \textbf{200} & \textbf{500} & \textbf{100} & \textbf{200} & \textbf{500}  & \textbf{100} & \textbf{200} & \textbf{500}  & \textbf{100} & \textbf{200} & \textbf{500}                           \\ \midrule 
Gold                                                 & 52.89             & 66.53             & 70.43                    & 16.37             & 27.7             & \ul{61.46}                       & 14.83            & 21.3           & 27.8                     & \ul{47.46}            & \ul{54.38}          &  59.41                  \\
LwTR                                            & 65.48             & \ul{73.24}   & \ul{81.45}                   & \ul{46.18}          & \ul{51.47}          & 54.87  &  \ul{21.59}            &  \ul{26.25}          & \ul{30.56}                      & 46.93             &  54.29        & \ul{59.76}                     \\
DAGA                                                  & 53.91             & 51.63             &  54.68                       & 33.29             & 43.07             & 54.64                  & 10.97            & 14.89             &  18.90                     &  34.67            &  41.98            & 48.72                         \\
MELM                                                 & 56.89             & 62.23             &  79.05                        & 11.94             & 31.55             & 45.68                         & 18.29             &  22.01            & 25.12 & 40.86         & 51.32       & 55.79              \\
GENIUS                                               & \ul{67.85}             & 58.2             & 80.36                     & 25.08             & 23.29             & 22.14                     & 20.08      & 16.87            & 21.41                    & 43.41         & 52.01         & 56.65                  \\
\textbf{CoDa \textit{(ours)}}                                & \cellcolor{magenta!20}\textbf{70.45}             & \cellcolor{magenta!20}\textbf{80.43}             & \cellcolor{magenta!20}\textbf{84.23}              & \cellcolor{magenta!20}\textbf{48.19}             & \cellcolor{magenta!20}\textbf{53.81}             & \cellcolor{magenta!20}\textbf{62.78}                       & \cellcolor{magenta!20}\textbf{23.22}             &        \cellcolor{magenta!20}\textbf{27.12}  &  \cellcolor{magenta!20}\textbf{32.45}                      &  \cellcolor{magenta!20}\textbf{49.56}            & \cellcolor{magenta!20}\textbf{54.85}             &  \cellcolor{magenta!20}\textbf{61.11}     \\ 
& $\pm0.91$           & $\pm0.84$             & $\pm0.91$ & $\pm0.45$ & $\pm0.65$ & $\pm0.72$ & $\pm0.49$              & $\pm0.79$  & $\pm0.34$          & $\pm0.54$             & $\pm0.12$            &  $\pm0.42$ \\
\bottomrule           
\end{tabular}
}
}
\caption{\small Result comparison for NER. CoDa outperforms baselines by 0.47\% - 7.19\%.}
\label{tab:ner}
\end{minipage}\hspace{0.02\textwidth}
\begin{minipage}{0.39\textwidth}
\scriptsize
\centering
\resizebox{\textwidth}{!}{
    \setlength{\tabcolsep}{1.8pt}
    {\renewcommand{\arraystretch}{1.05}%
    \begin{tabular}{l|ccc|ccc}
        \toprule
        \multicolumn{1}{c|}{\multirow{2}{*}{\textbf{Model}}} &  \multicolumn{3}{c|}{\textbf{SQuAD}}                     & \multicolumn{3}{c}{\textbf{NewsQA}}\\  
        \multicolumn{1}{c|}{} & \textbf{100} & \textbf{200} & \textbf{500} & \textbf{100} & \textbf{200} & \textbf{500}                           \\ \midrule 
        Gold  & 11.64                         & 19.71                         & 26.32          &  22.45             & 30.14             & 45.65  \\
        BackTrans & 17.47             & 22.60             & 29.07                  & 27.32             & 34.98             & 47.21  \\
        EDA  & 17.07                      & 22.39                         & 28.98                     & 29.31              & 35.81             & 49.90  \\
         AEDA & 17.95             & 23.50             & 29.20                    & 29.87              & 36.80             & 50.24   \\
        SSMBA & 16.97            & 22.27             & 28.51                    & 28.89             & 33.27             &  47.56  \\
        GENIUS & \ul{33.15}             & \ul{42.65}             & \ul{56.52}                    & \ul{38.88}             &  \ul{47.36}            & \ul{57.32}   \\
        \textbf{CoDa \textit{(ours)}} &  \cellcolor{magenta!20}\textbf{36.21}             &  \cellcolor{magenta!20}\textbf{44.89}             &  \cellcolor{magenta!20}\textbf{57.90}                    & \cellcolor{magenta!20}\textbf{39.98}             & \cellcolor{magenta!20}\textbf{49.86}             & \cellcolor{magenta!20}\textbf{58.94} \\     
        & $\pm0.21$           & $\pm0.34$             & $\pm0.11$ & $\pm0.35$ & $\pm0.15$ & $\pm0.22$ \\
        \bottomrule
        \end{tabular}
    }
}
\caption{\small Result comparison for QA. CoDa outperforms baselines by 1.10\% - 3.06\%.}
\label{tab:qa}
\end{minipage}
\end{table*}

\subsection{Constructing the Instruction}
\label{sec:instruction}
After extracting the constraints from $d$, we verbalize the constraints to a single instruction for prompting an instruction-tuned LLM. The verbalization is done through fixed hand-written templates. An example of an instruction is shown in Fig.~\ref{fig:main_diag}.

\section{Experimental Setup}
\label{sec:experimentl_setup}

{\noindent \textbf{Baselines.}} Gold-only refers to training our model only on the low-resource gold data. For sequence classification (SC), we compare CoDa with text editing baselines: EDA \cite{wei2019eda}, AEDA \cite{karimi-etal-2021-aeda-easier}, and AMR-DA~\cite{shou-etal-2022-amr}, learning-based infilling baselines: SSMBA \cite{ng-etal-2020-ssmba}, GENIUS(-\textbf{ft}) \cite{guo2022genius}, PromDA~\cite{wang-etal-2022-promda}, LLM-based prompting baselines: ZeroGen~\cite{ye-etal-2022-zerogen}, GPT3Mix~\cite{yoo-etal-2021-gpt3mix-leveraging} and rephrasing baselines: BackTrans~\cite{yu2018qanet}. For the Intent Classification task, specifically in SC, we add another LLM-based prompting baseline: PromptMix~\cite{sahu2023promptmix}. For Named Entity Recognition (NER), we compare CoDa with LwTR \cite{dai-adel-2020-analysis}, DAGA~\cite{ding-etal-2020-daga}, MELM \cite{zhou2021melm}, PromDA~\cite{wang-etal-2022-promda} and ACLM~\cite{ghosh-2023-aclm}. Finally, for question answering (QA), we compare it with ZeroGen, BackTrans, GENIUS, EDA, and AEDA. Details on the working of all baselines are provided in Section~\ref{sec:baseline_details}.
\vspace{0.5mm}

{\noindent \textbf{Datasets.}} To demonstrate CoDa's flexibility, we evaluate it across various challenging datasets belonging to a wide range of domains. For SC, we employ Huffpost~\cite{huffpost} (news category classification), Yahoo~\cite{zhang2015character} (answer topic classification), OTS~\cite{drawzeski-etal-2021-corpus} (\textit{legal} online service unfairness level classification), ATIS~\cite{coucke2018snips} and Massive~\cite{fitzgerald2022massive} (Intent Classification). For NER, we employ ConLL-2003~\cite{tjong-kim-sang-de-meulder-2003-introduction},  OntoNotes-5.0~\cite{pradhan2013towards} (news domain), EBMNLP~\cite{nye2018corpus} and BC2GM~\cite{krallinger2015chemdner} (bio-medical). Finally, for QA, we employ SQuAD~\cite{rajpurkar2016squad} and NewsQA~\cite{trischler-etal-2017-newsqa}. Details on each dataset and dataset statistics are provided in Section~\ref{sec:dataset_details}.
\vspace{0.5mm}


\begin{figure*}
    \includegraphics[width=1\textwidth]{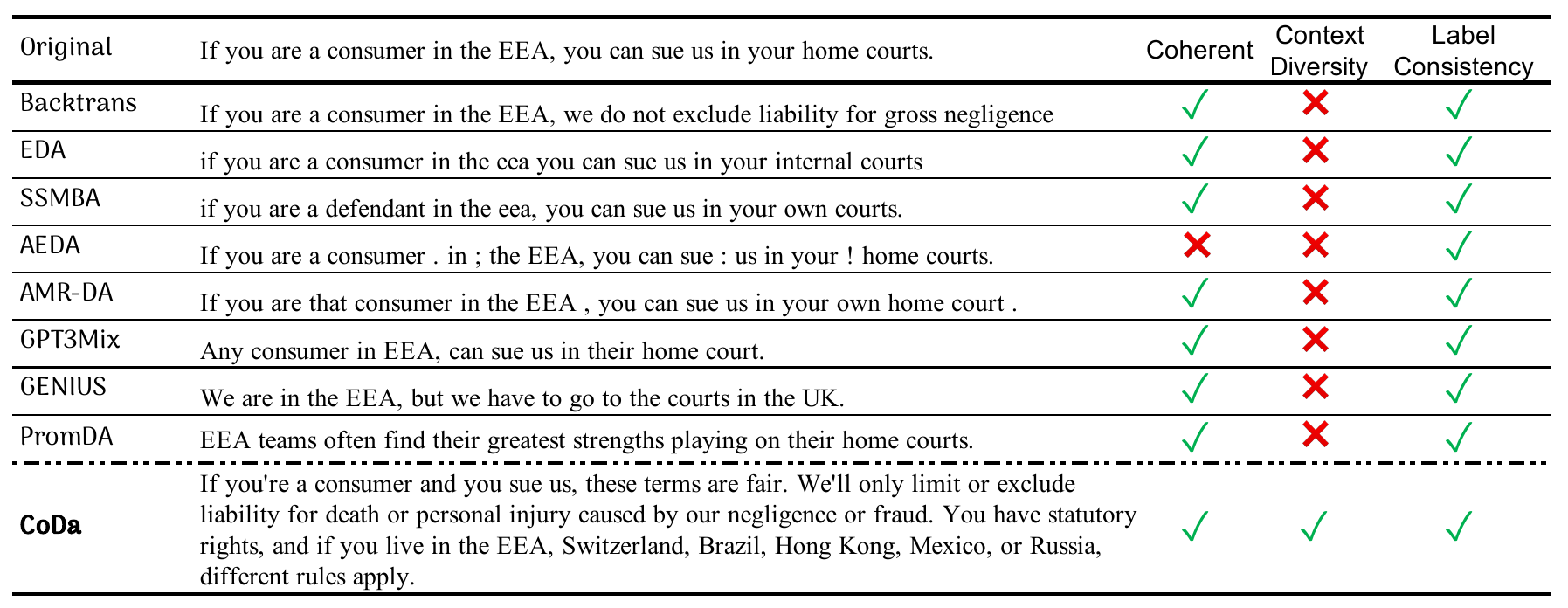}
    \caption{Augmentation examples on the OTS dataset. All generations are produced in a low-resource setting (500 training examples). CoDa generates augmentations that are coherent, diverse, and label-consistent.}
    \label{fig:ots}
\end{figure*}

{\noindent \textbf{Hyper-parameter settings.}} We prompt LLama-13B with a temperature of 0.5, top-\textit{p} of 1.0, top-\textit{k}=50. For all downstream NLU tasks, we employ BERT\textsubscript{base-uncased}~\cite{devlin2019bert} as our encoder (except OTS where we employ legal-longformer\textsubscript{large}~\cite{chalkidis-garneau-etal-2023-lexlms}). We fine-tuned our encoder with a batch size of 4,8 for 100 and 200 splits and 16 for 500 and 1000 splits.  For NER specifically, we employ the flair library~\cite{akbik2019flair} with an initial lr of $1e^{-5}$ and constant decay. Appendix~\ref{sec:hyper-parameter} provides hyper-parameter tuning experiments. We report the micro-average F$_1$ score averaged across 3 runs for 3 random seeds.



\section{Results and Analysis}
\label{sec:results_analysis}

{\noindent \textbf{Quantitative Analysis.}} Table \ref{tab:classification}, \ref{tab:ner}, and \ref{tab:qa} compared CoDa with all our baselines on the tasks of SC, NER, and QA, respectively. CoDa outperforms our baselines in SC by 0.12\% - 5.94\%, NER by 0.47\% - 7.19\%, and QA by 1.10\% - 3.06\%. Though most prior methods proposed for one domain generally underperform in the other~\cite{ghosh-etal-2023-dale}, CoDa consistently outperforms these methods in all domains with varying semantic and syntactic properties, emphasizing its domain-agnostic nature.

{\noindent \textbf{Qualitative Analysis.}} Table \ref{tab:qual_table} compares the generation quality of CoDa with all our baselines (averaged baseline-wise across all tasks and splits) on the measures of perplexity \cite{jelinek1977perplexity}, diversity (average number of new tokens introduced in $R$ augmentations)  and length diversity (average absolute difference in length of source and $R$ augmentations). CoDa outperforms most of our baselines in all settings. Additionally, as observed in Table~\ref{tab:qual_table}, unlike other learning-based methods in literature, the diversity of augmentations by CoDa does not depend on the number of gold training samples available. It performs equally well in both 100 and 500 splits.

Fig.~\ref{fig:ots} compares CoDa augmentations with other baselines in literature with a gold training sample taken from the OTS dataset. Generating augmentation on the OTS dataset, which belongs to the legal domain, is inherently difficult due to the formalized nature of legal language~\cite{ghosh-etal-2023-dale}. As we can see, CoDa generates augmentations that are coherent, diverse, and label-consistent. More examples are provided in Fig.~\ref{fig:atis}, ~\ref{fig:conll} and ~\ref{fig:squad}. Additionally, Appendix~\ref{sec:faithfulness} evaluates how faithful LLaMa-2 was in following the constraints in the instructions.

\begin{table}[t]
\small
\centering
\resizebox{\columnwidth}{!}{%
\begin{tabular}{l|cc||cc}
\toprule \midrule
\textbf{Method}     & \textbf{Perplexity($\downarrow$)} & \textbf{Diversity($\uparrow$)}  & \textbf{Perplexity($\downarrow$)} & \textbf{Diversity($\uparrow$)}\\ \midrule
                    & \multicolumn{2}{c||}{100}                                        & \multicolumn{2}{c}{500}        \\ \midrule
EDA                           & 104.93                             & 115.89                                                   &  118.83                            & 156.21                                                   \\
GENIUS      & \ul{24.90}                           & 120.64                         &  \ul{25.43}                 & 126.32                          \\
GPT3Mix                         & 88.77                             & \ul{146.89}                          &  75.17                          &  \ul{163.32}                         \\
BackTrans                          & 240.93                             & 132.51                           &  74.91                      &  56.31                          \\
AMR-DA                        & 61.59                             & 77.94                          &  50.73                         & 84.81                                          \\
LwTR                     & 135.89                             & 94.77                         &  139.93                      & 99.63                                      \\
\textbf{CoDa} \textit{(ours)}                      & \cellcolor{magenta!20}\textbf{22.44}                       & \cellcolor{magenta!20}\textbf{152.34}                         &                    \cellcolor{magenta!20}\textbf{23.33}       &  \cellcolor{magenta!20}\textbf{165.81}  \\ \midrule \bottomrule
\end{tabular}%
}
\caption{\small Quantitative evaluation of generation quality on the measures of perplexity and token diversity. CoDa outperforms all our baselines on all metrics.
}
\vspace{-1.5em}
\label{tab:qual_table}
\end{table}

\section{Conclusion}
\label{sec:conclusion}

We present CoDa, a simple and controllable data augmentation technique for low-resource NLP. CoDa extracts simple heuristic-based constraints from source sentences and verbalizes them to construct and instruction, which is then used to prompt LLMs to generate augmentations. CoDa is \textit{training-free} and works with any out-of-the-box instruction-tuned LLM. Beyond providing explicit control, CoDa is also flexible, i.e., constraints can be easily replaced or added, enhancing its suitability across diverse domains.

\section*{Limitations and Future Work}
\label{sec:limitations}
Despite its effectiveness, CoDa suffers from various limitations, which we would like to mention. These limitations will remain our primary focus in future work. The limitations are as follows:

\begin{itemize}
    \item LLMs often struggle to follow complex constraints in the instruction for text generation~\cite{lu2023bounding}. We overcome this problem in CoDa by employing simple constraints. However, we acknowledge that data augmentation for complex domains and tasks may need to employ more complex constraints. Thus, as part of future work, we would like to employ recent advances in compositional prompting for breaking down complex constraints into simpler instructions.
    \item Although being \textit{training free}, CoDa is computationally more expensive during inference time compared to prior art as it employs LLMs. As also shown in Section \ref{sec:choice}, the overall performance of CoDa takes a slight hit when LLaMa-7B was employed instead of the 13B version. However, we acknowledge that as smaller models get better at following instructions, CoDa can perform more efficiently.
\end{itemize}


\bibliography{anthology,custom}

\appendix
\section{Hyper-parameter Tuning}
\label{sec:hyper-parameter}


\subsection{Effect of augmentation rounds $R$}
\label{sec:augrounds}

Table \ref{tab:aug} compares the performance of CoDa at different values of $R$. Augmenting the training dataset with several augmentation rounds $R$ proves effective until the model overfits to the training data. The observation is similar to prior work in data augmentation for NLU tasks \cite{zhou2021melm,ghosh-2023-aclm}.

\begin{table}[h]
    \centering
    \small
    \resizebox{\columnwidth}{!}{
    \begin{tabular}{c|ccccccc}
    \hline
         \textbf{$R$} & \textbf{1} & \textbf{2} & \textbf{3} & \textbf{4} & \textbf{5} & \textbf{6} & \textbf{7}\\
        \hline
          $F_1$& 61.74 & 62.05 & 62.31 & 63.01 & \cellcolor{magenta!20}\textbf{63.16} & 62.99 & 61.23  \\
         \hline
    \end{tabular}
    }
    \caption{F1 for various settings of $R$. All values are averaged across all datasets for all low-resource settings.}
    \label{tab:aug}
\end{table}

\subsection{Choice of LLM}
\label{sec:choice}

Table~\ref{tab:llm} compares the performance of CoDa employing different open-source LLMs. Beyond LLaMa-13B employed in our paper, we also compare performance with Mistral-7B~\cite{jiang2023mistral} and LLaMa-7B. As we see, employing LLaMa-7B takes a hit of 0.18\% on the final performance, while employing Mistral-7B takes a hit of 1.44\% on the final performance. We also noticed several instances of hallucination with Mistral-7B, where the output of the LLM was completely different from the given instruction. This was not the case with the LLaMa family of models, and performance generally improved with a larger model owing to a better quality of generations and better abilities to follow instructions.

\begin{table}[h]
    \centering
    \small
    \begin{tabular}{|c|c|}
    \hline
         \textbf{LLM} & \textbf{F1-Micro}\\
         \hline
          Mistral-7B & 61.72 \\
          LLaMa-7B & \ul{ 62.98} \\
          LLaMa-13B & \cellcolor{magenta!20}\textbf{63.16} \\
         \hline
    \end{tabular}
    \caption{F1 micro averaged across tasks for various LLMs.}
    \label{tab:llm}
\end{table}

\section{Faithfulness in following instruction constraints}
\label{sec:faithfulness}

Table~\ref{tab:constraints} illustrates the accuracy of augmentations produced by LLaMa-13B in adhering to the constraints specified in the instruction. We only illustrate accuracies of Lexical and Length constraints as they are easily quantifiable. Other constraints require human evaluation, which remains part of future work. We report the accuracy as our metric for faithfulness, wherein we consider a generation as accurate for the constraint if it completely follows the constraint, else inaccurate. Additionally, we also report a 75\% threshold for both the constraints, whereby we consider the generation as accurate if it follows 75\% of the constraint (e.g.,75\% of the total keywords mentioned are in generation and the total tokens in the generation lie between 75\% of the maximum and minimum lengths). Although LLaMa-13B demonstrates moderate proficiency in adhering to constraints, the anticipated improvement in instruction-following capabilities of LLMs is likely to enhance these metrics further. Furthermore, the fact that CoDa surpasses the performance of many existing models in the literature, despite its moderate ability to follow constraints, suggests a significant promise for CoDa as an augmentation generation scheme when integrated with more advanced LLMs. 

An observed trend is that models demonstrate strong performance on familiar datasets such as CoNLL-2003, potentially due to these datasets being included in their pre-training corpus. Additionally, our models exhibit improved performance under 75\% threshold constraints. This suggests a balance must be struck between the creative output and adherence to constraints in LLM generations. Although creativity is crucial for generating diverse augmentations, following constraints is key for maintaining consistency. In future work, we aim to investigate more effective methods for balancing this trade-off.

\begin{table}[h]
    \centering
    \large
    \resizebox{\columnwidth}{!}{
        \begin{tabular}{|c|c|c|c|c|}
        \hline
             \textbf{Task} & \textbf{Lexical} & \textbf{Lexical 75\%} & \textbf{Length} & \textbf{Length 75\%}\\
             \hline
              HuffPost & 24.64 & 26.09 & 51.31 & 55.02\\
              Yahoo &27.28 & 28.12& 51.06 & 54.48\\
              OTS & 21.83 & 23.32 & 50.98 & 53.95\\
              ATIS & 41.1 & 43.5 &50.2 & 51.52\\
              MASSIVE & 26.26 & 28.32 &50.22 &51.52\\
              CoNLL-2003 & 67.72 & 73.31 &51.13 &53.82 \\
              OntoNotes & 36.33 & 48.7 &50.59 & 53.12\\
              EBMNLP & 41.05 & 45.46 &50.72 & 53.17\\
              BC2GM & 41.45 &  48.82 &50.6 & 53.17\\
              SQUAD & 32.56 &  40.87 &52.12 & 55.82\\
              NEWSQA & 33.45 &  42.18 &51.98 & 54.87\\
             \hline
        \end{tabular}
    }
    \caption{\small Faithfulness of generated augmentations. Scores reported correspond to average accuracy, where we attribute an augmentation as accurate if it perfectly follows the constraint in the given instruction; otherwise, we attribute it as inaccurate.}
    \label{tab:constraints}
\end{table}

\section{Dataset Details}
\label{sec:dataset_details}

\begin{table*}[t]
    \centering
    \resizebox{\textwidth}{!}{
    \begin{tabular}{llclcc}
    \toprule
         \textbf{Dataset} & \textbf{Source} & \textbf{Sub-domain} & \textbf{Task Type} & \textbf{Training/Dev/Test Instances} & \textbf{Classes} \\
         \midrule
          HuffPost &\citet{huffpost} & HuffPost website & Multi-class classification & 67490/16891/16891 & 5\\
         Yahoo & \citet{zhang2015character} & Yahoo Answers  & Multi-class classification & 1375404/58966/58966 & 10  \\
         OTS-UL & \citet{drawzeski-etal-2021-corpus} & EU Law & Multi-class classification & 2074/191/417 & 3 \\
         ATIS & \citet{CNTK_2023} & Travel enquiry & Intent Classification & 4972/888/888 & 17 \\
         MASSIVE & \citet{fitzgerald2022massive} & Diverse & Intent Classification & 11500/2030/2970 & 60 \\
         CoNLL-2003 & \citet{tjong-kim-sang-de-meulder-2003-introduction}  & English news articles & Named Entity Recognition & 14041/3250/3453 & 4 \\
         OntoNotes-5.0 & \citet{pradhan2013towards} & Diverse & Named Entity Recognition & 115812/15680/12217 & 36 \\
         BC2GM & \citet{krallinger2015chemdner} & Biomedical & Named Entity Recognition & 15197/3061/6325 & 2 \\
         EBMNLP & \citet{nye2018corpus} & Biomedical & Named Entity Recognition & 35005/10123/6193 & 7 \\
         SQUAD & \cite{rajpurkar2016squad} & Wikipedia Articles & Question Answering & 87600/10600/- & - \\
         NEWSQA & \cite{trischler-etal-2017-newsqa} & CNN Articles & Question Answering & 92549/5126/5166 & - \\
         \bottomrule
    \end{tabular}
    }
    \vspace{-2mm}
    \caption{Statistics for each downstream NLU datasets used in our experiments. As described in Section \ref{sec:experimentl_setup}, we derive low-resource splits from these original datasets for our experiments.}
    \label{tab:nlu_datasets}
    \vspace{-5mm}
\end{table*}

\subsection{Classification}
\label{subsec:classification}

{\noindent \textbf{HuffPost.}}  The HuffPost dataset \cite{huffpost} is a popular multiclass classification dataset in NLP. It is a collection of news articles from the HuffPost website, covering a wide range of topics, including politics, business, entertainment, and more. For multiclass classification, the HuffPost dataset is labeled with a diverse set of categories and for our experiments, we take sentences from five categories, including politics, sports, entertainment, tech, and business. Dataset statistics can be found in Table \ref{tab:nlu_datasets}.
\vspace{0.5mm}

{\noindent \textbf{Yahoo.}} The Yahoo Answers topic classification dataset \cite{zhang2015character} is a widely used dataset for multi-class text classification tasks. It is derived from the Yahoo Answers community-driven question-answering platform, where users ask questions on various topics, and community members provide answers. The dataset contains a large number of question-and-answer pairs covering a wide range of categories or topics. Each question in the dataset is associated with one primary category. The primary categories span diverse subjects, including Society \& Culture, Science \& Mathematics, Health, Education \& Reference, Computers \& Internet, Sports, Business \& Finance, Entertainment \& Music, Family \& Relationships, Politics \& Government, Travel, Cars \& Transportation, Food \& Drink, Games \& Recreation, Home \& Garden, Local Businesses, News \& Events, Pets, Beauty \& Style and Pregnancy \& Parenting. Dataset statistics can be found in Table \ref{tab:nlu_datasets}.

{\noindent \textbf{OTS-UL.}} Online Terms of Service (OTS) \cite{drawzeski-etal-2021-corpus} attempt to automatically detect unfair clauses in Terms of Service. The input to the model is a sentence, and the output presents the sentence classified into three
levels of unfairness. The dataset setting used in our paper is similar to \citep{niklaus2023lextreme}. Dataset statistics can be found in Table \ref{tab:nlu_datasets}.

\subsection{Named Entity Recognition}
\label{subsec:ner}

{\noindent \textbf{CoNLL-2003.}} The CoNLL-2003 dataset \cite{tjong-kim-sang-de-meulder-2003-introduction} is a widely used benchmark dataset for Named Entity Recognition (NER) tasks in NLP. It was created for the Conference on Computational Natural Language Learning (CoNLL) shared task in 2003. The dataset consists of news articles from the Reuters Corpus, a collection of English news articles. It is annotated with four named entities: person, organization, location, and miscellaneous entities (such as dates and percentages). The annotations indicate the boundaries of the named entities within the text. Dataset statistics can be found in Table \ref{tab:nlu_datasets}.
\vspace{0.5mm}

{\noindent \textbf{Ontonotes 5.0.}} Ontonotes 5.0 \citet{pradhan2013towards} is a widely used dataset in the field of Natural Language Processing (NLP) and specifically for Named Entity Recognition (NER) tasks. It is a large-scale corpus that provides annotations for a variety of linguistic phenomena, including named entities, across multiple languages. The dataset contains a diverse range of text genres, including news articles, conversational data, and web data, making it suitable for training and evaluating NER models in different domains. It covers three languages: English, Chinese, and Arabic. The dataset is annotated with 11 categories: Person, Organization, Location, Date, Time, Money, Percent, Quantity, Ordinal and Miscellaneous. Dataset statistics can be found in Table \ref{tab:nlu_datasets}.
\vspace{0.5mm}

{\noindent \textbf{EBMNLP.}} EBMNLP \citet{nye2018corpus} is a widely used dataset in the field of Biomedical Named Entity Recognition (BioNER) tasks. It is a corpus of richly expert-annotated abstracts of medical articles describing clinical randomized controlled trials. The dataset facilitates easy search and organization of published literature on randomized controlled trials, addressing the current challenges impeding the goals of evidence-based medicine (EBM). The dataset is annotated with 3 categories: Outcome, Intervention and Participant. Dataset statistics can be found in Table \ref{tab:nlu_datasets}.
\vspace{0.5mm}

{\noindent \textbf{BC2GM.}} BC2GM \citet{krallinger2015chemdner} is a widely used dataset in the field of Biomedical Named Entity Recognition (BioNER) tasks. This dataset is a part of the CHEMDNER large scale corpus which includes annotation of chemical entities as well as named entities in the biomedical and other domains. The dataset is annotated with 1 categoriy: Gene. Dataset statistics can be found in Table \ref{tab:nlu_datasets}.

\subsection{Intent Classification}
\label{subsec:intent}
{\noindent \textbf{ATIS.}} The ATIS (Airline Travel Information System) dataset\footnote{\url{https://github.com/howl-anderson/ATIS_dataset/tree/master}} is a widely used benchmark dataset for intent classification in the field of NLU. It was developed to address understanding user intents in the context of airline travel information. The dataset consists of queries or utterances that users might input when interacting with a flight reservation system. Each query is labeled with an intent representing the user's intention or purpose behind the query. 
The dataset is labeled with intents that are: Flight-Booking, Flight-Status, Flight-Information, Ground-Service, Airfare, Airport-Information, Travel-Preferences, Flight-Cancellation, and None/No-Intent. Dataset statistics can be found in Table \ref{tab:nlu_datasets}.
\vspace{0.5mm}

{\noindent \textbf{MASSIVE.}} The MASSIVE (Multilingual Amazon Slu resource package for Slot-filling) \citet{fitzgerald2022massive} dataset is a widely used benchmark dataset for intent classification in the field of NLU. It contains 1M realistic, parallel, labeled virtual assistant utterances spanning 51 languages, 18 domains, 60 intents, and 55 slots. The dataset is labeled with intents some of which are: Alarm set, Play music, Audio volume mute, Weather query, Takeaway order and General joke etc. Dataset statistics can be found in Table \ref{tab:nlu_datasets}.

\subsection{Question Answering}
\label{subsec:qa}
{\noindent \textbf{SQUAD.}} The SQUAD (Stanford Question Answering Dataset) \cite{rajpurkar2016squad} is a reading comprehension dataset, consisting of questions posed by crowdworkers on a set of Wikipedia articles, where the answer to every question is a segment of text, or span, from the corresponding reading passage, or the question might be unanswerable. Dataset statistics can be found in Table \ref{tab:nlu_datasets}.
\vspace{0.5mm}

{\noindent \textbf{NEWSQA.}} NewsQA (News Question Answering) \cite{trischler-etal-2017-newsqa} is a challenging machine comprehension dataset of over 100,000 human-generated question-answer pairs. Crowdworkers supply questions and answers based on a set of over 10,000 news articles from CNN, with answers consisting of spans of text from the corresponding articles. Dataset statistics can be found in Table \ref{tab:nlu_datasets}.

\section{Baseline Details}
\label{sec:baseline_details}


{\noindent \textbf{SSMBA.}} SSMBA \cite{ng-etal-2020-ssmba} generates synthetic training examples by using a pair of corruption and reconstruction functions to move randomly on a data manifold.
\vspace{0.5mm}

{\noindent \textbf{AEDA.}} AEDA \cite{karimi-etal-2021-aeda-easier} is similar to EDA but only employs random insertion of punctuation marks in the original text to generate synthetic augmentations.
\vspace{0.5mm}

\begin{table*}[t]
    \resizebox{\textwidth}{!}{
    \begin{tabular}{|l|l|}
    \toprule
         \textbf{Dataset} & \textbf{Syntactic Examples} \\
         \midrule
         OTS & 1. \textbf{Constraint}: ADV PRON AUX PUNCT ADP PRON NOUN PUNCT VERB NOUN ADP DET NOUN PUNCT\\&\textbf{Generation 1}: Quickly he can, upon her request, examine the document.\\&\textbf{Generation 2}: Additionally, he shall submit the document to the court.\\
         & 2. \textbf{Constraint}: NOUN AUX PART ADJ ADP NOUN PRON AUX PART VERB PRON NOUN PUNCT\\&\textbf{Generation 1}: Contractors have, under new regulations, completed their work.\\&\textbf{Generation 2}: Judge may have been impartial in the legal proceedings.\\
         \midrule
         EBMNLP & 1. \textbf{Constraint}: DET ADJ NOUN AUX VERB ADP NOUN NOUN PUNCT\\&\textbf{Generation 1}: The molecular analysis revealed a genetic mutation in the patient.\\&\textbf{Generation 2}: The experimental procedure was conducted on laboratory samples.\\
         & 2. \textbf{Constraint}: NOUN NOUN NOUN ADP NUM NOUN AUX VERB VERB ADP NUM NOUN NOUN PUNCT\\&\textbf{Generation 1}: EXPERIMENT A cohort of 50 samples was collected from 3 laboratory facilities.\\&\textbf{Generation 2}: STUDY A group of 100 patients underwent testing in two medical centers.\\
         \midrule
         BC2GM & 1. \textbf{Constraint}: PROPN PROPN PROPN ADP DET NOUN ADP NOUN NOUN ADP NOUN PUNCT\\&\textbf{Generation 1}: Polymerase chain reaction for the detection of genetic mutations in patients.\\&\textbf{Generation 2}: Hormone receptor status in the evaluation of breast cancer in women.\\
         & 2. \textbf{Constraint}: NOUN ADP NOUN NOUN CCONJ NOUN PUNCT\\&\textbf{Generation 1}: Analysis of protein structures and 3,4-dihydroxyphenylalanine.\\&\textbf{Generation 2}: Exploration of biochemical pathways and 2,3-dimethylbutane.\\
         \bottomrule
    \end{tabular}
    }
    \vspace{-2mm}
    \caption{\small Examples of a couple of documents corresponding to a single POS sequence in formal domains like legal (OTS) and bio-medical (EBMNLP and BC2GM). We emphasize that syntactic constraints help generate augmentations better aligned to the domain.}
    \label{tab:syntactic_constraints}
    \vspace{-5mm}
\end{table*}

{\noindent \textbf{GENIUS.}} GENIUS \cite{guo2022genius}, pre-trains and optionally fine-tunes BART \cite{lewis2019bart} on a denoising objective using sketches generated with an extreme masking algorithm. The extreme masking algorithm just preserves keywords in a sentence and masks everything else. 
\vspace{0.5mm}

{\noindent \textbf{MELM.}} MELM \cite{zhou2021melm}, which stands for Masked Entity Language Modeling, suggests the fine-tuning of a transformer-encoder-based PLM on linearized labeled sequences through masked language modeling. In low-resource scenarios, MELM surpasses all other baselines and prior techniques on the CoNLL 2003 NER dataset across four languages, including mono-lingual, cross-lingual, and multi-lingual settings.
\vspace{0.5mm}

{\noindent \textbf{DAGA.}} DAGA \cite{ding-etal-2020-daga}, short for Data Augmentation with a Generation Approach, suggests the training of a one-layer LSTM-based recurrent neural network language model (RNNLM) by maximizing the probability of predicting the next token using linearized sentences. For sentence generation, they employ random sampling to create entirely new sentences, with the model being fed only the $\lbrack\textbf{BOS}\rbrack$ token.
\vspace{0.5mm}


{\noindent \textbf{LwTR.}} LwTR \cite{dai-adel-2020-analysis} replaces a token in a sentence with another token of the same label; the token is randomly selected from the training set.

{\noindent \textbf{PromDA.}} PromDA \cite{wang-etal-2022-promda} proposes a data augmentation framework based on T5 that trains soft prompts using a novel keyword-to-sentence algorithm.

{\noindent \textbf{AMR-DA.}} AMR-DA~\cite{shou-etal-2022-amr} converts a sample document from a dataset to an AMR graph, modifies the graph according to various data augmentation policies, and then generates augmentations from graphs. The method combines both sentence-level techniques like back translation and token-level techniques like EDA.

{\noindent \textbf{PromptMix.}} PromptMix~\cite{sahu2023promptmix} PromptMix prompts instruction-tuned LLMs to generate augmentations for text classification tasks that are close to the class boundary.

{\noindent \textbf{ZeroGen.}} ZeroGen~\cite{ye-etal-2022-zerogen}, similar to PromptMix, generates data using LLMs but in a zero-shot manner without any gold data. It prompts pre-trained LLMs (not instruction fine-tuned) for data synthesis.

We do not consider more recent baselines provided by \citet{cai-etal-2023-graph}, \citet{hu2023entitytotext} and \citet{rahamim-etal-2023-text} as the code for the same was not available at the time of writing the paper. Additionally, we do not consider \citet{zhou-etal-2022-flipda} as label flipping is not applicable for our paper for all tasks considered, and \citet{chen-etal-2022-style} as style transfer is better suited for cross-domain tasks and applying it to single domain tasks is not trivial. Finally, we do not consider \citet{yu2023large} as it requires manual human intervention for attribute extraction for a dataset.
\section{Additional Details}
\label{sec:addition_details}

\subsection{Examples of syntactic constraints in formal domains}
\label{sec:syntac}

Table~\ref{tab:syntactic_constraints} provides examples of documents from domains with formal language, like legal and bio-medical. Each example provides two corresponding documents to a POS sequence, emphasizing that syntactic constraints help generate augmentations better aligned to the domain in formal domains.

\subsection{Qualitative Examples}
\label{sec:qual}

\begin{figure*}
    \includegraphics[width=1\textwidth]{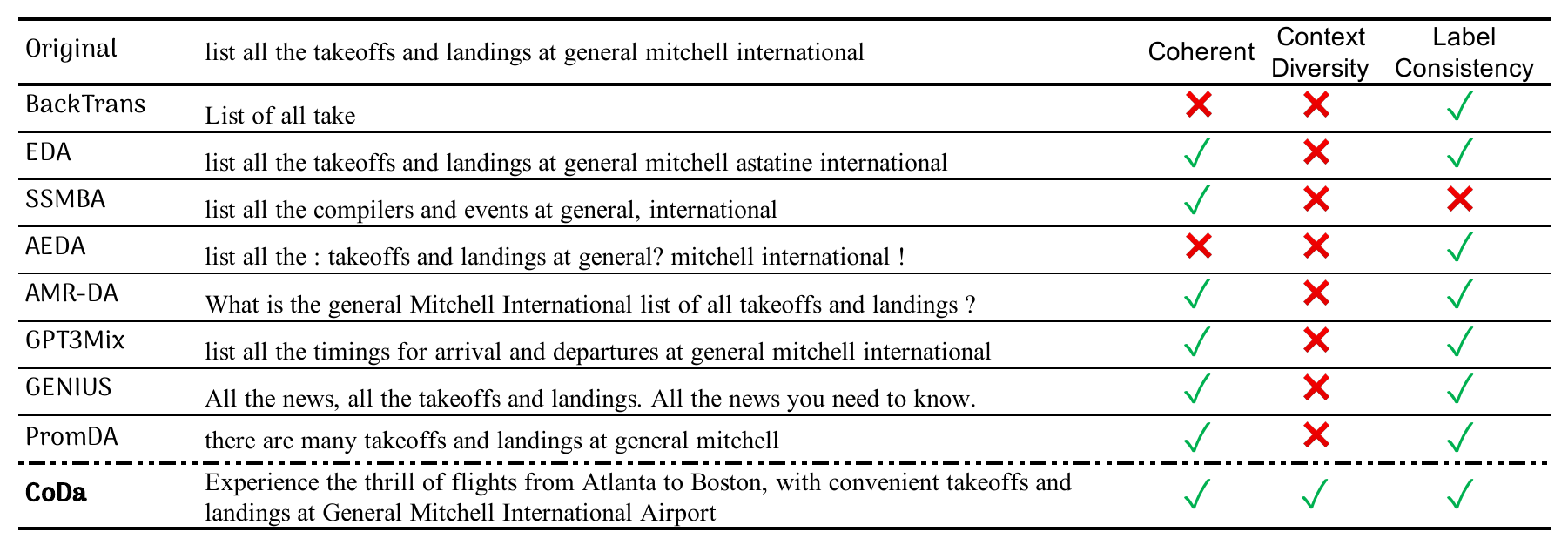}
    \caption{\small{Augmentation examples on the ATIS dataset. All generations are produced in a low-resource setting (500 training examples).}}
    \label{fig:atis}
\end{figure*}

\begin{figure*}
    \includegraphics[width=1\textwidth]{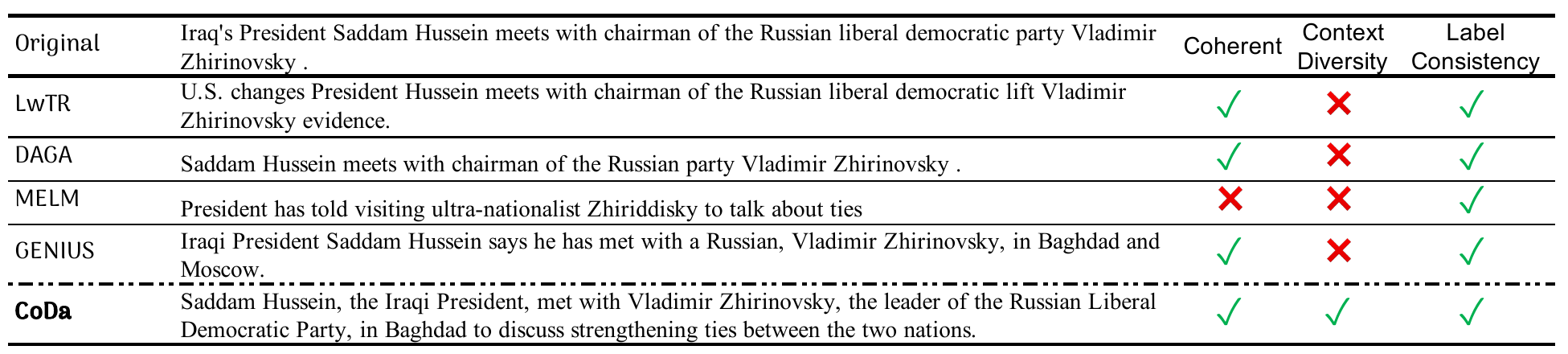}
    \caption{\small{Augmentation examples on the CoNLL-2003 dataset. All generations are produced in a low-resource setting (500 training examples).}}
    \label{fig:conll}
\end{figure*}

\begin{figure*}
    \includegraphics[width=1\textwidth]{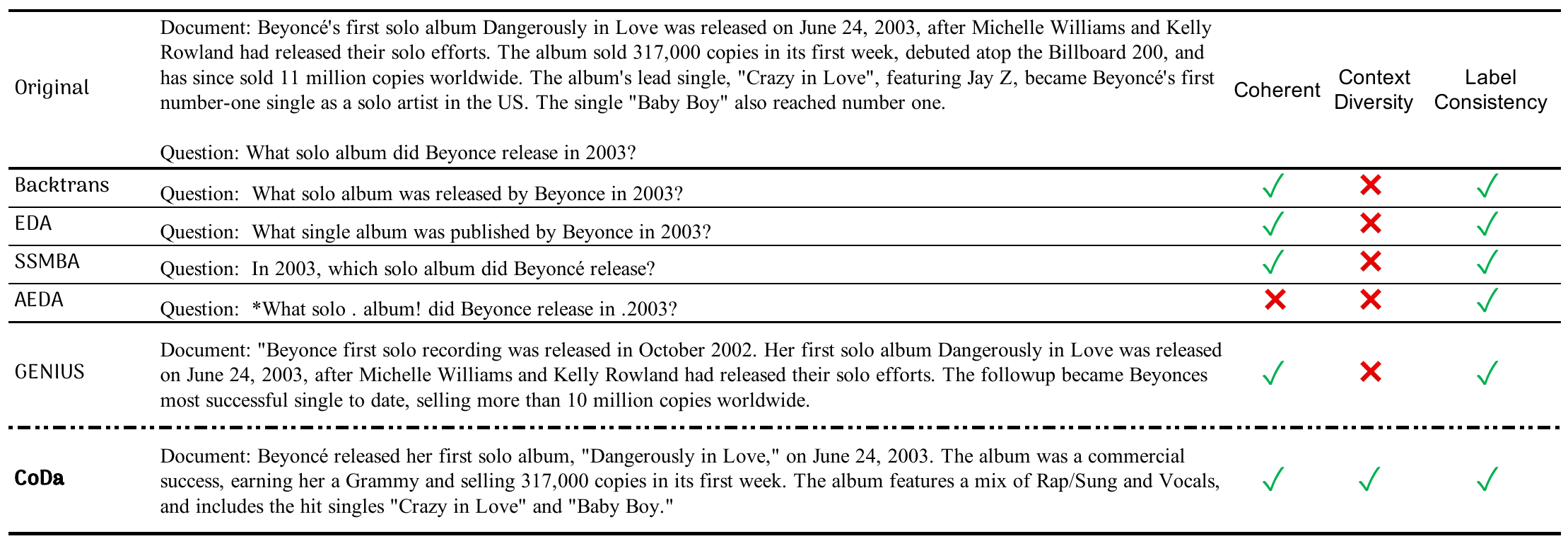}
    \caption{\small{Augmentation examples on the SQUAD dataset. All generations are produced in a low-resource setting (500 training examples).}}
    \label{fig:squad}
\end{figure*}

Fig.~\ref{fig:ots}, \ref{fig:atis}, \ref{fig:conll} and \ref{fig:squad} provide additional qualitative examples of augmentations generated using CoDa and compares them with other baselines. CoDa consistently generates more diverse and consistent augmentations over prior art.

\section{Extra Details}
\label{sec:extra_details}

\subsection{Model Parameters}
BERT\textsubscript{base} has $\approx$~110M 12-layers of encoder, 768-hidden-state, 2048 feed-forward hidden-state, and 8-heads. legal-longformer\textsubscript{large} has $\approx$ 149M 30 layers of encoder, 768-hidden-state, 3072 feed-forward hidden-state, and 12-heads. LLaMa-13B is a 13B parameter model and LLaMa-7B is a 7B parameter model.


\subsection{Compute Infrastructure}
All our experiments are conducted on NVIDIA A100 and NVIDIA A6000 GPUs. We batch prompted LLaMa-2 13B and LLaMa-2 7B, with a BS of 16, where LLaMa-2 performed distributed inference on 4 A6000 GPUs. Fine-tuning on the downstream tasks uses 4 A100 GPUs.

\subsection{Implementation Software and Packages}
We implement all our models in PyTorch\footnote{\href{https://pytorch.org}{https://pytorch.org/}} and use the HuggingFace\footnote{\href{https://huggingface.co/}{https://huggingface.co/}} implementations of BERT\textsubscript{base}, legal-longformer\textsubscript{large}, LLaMa-13B and LLaMa-7B. For NER specifically, we employ the Flair \footnote{\href{https://github.com/flairNLP/flair}{https://github.com/flairNLP/flair}} library.

We also use the following repositories for running the baselines: BackTrans~\cite{yu2018qanet}, EDA\footnote{\href{https://github.com/jasonwei20/eda_nlp}{https://github.com/jasonwei20/eda\_nlp}}\cite{wei2019eda}, AEDA\footnote{\href{https://github.com/akkarimi/aeda_nlp}{https://github.com/akkarimi/aeda\_nlp}}~\cite{karimi-etal-2021-aeda-easier}, AMR-DA\footnote{\href{https://github.com/zzshou/amr-data-augmentation}{https://github.com/zzshou/amr-data-augmentation}}~\cite{shou-etal-2022-amr}, SSMBA\footnote{\href{https://github.com/nng555/ssmba}{https://github.com/nng555/ssmba}}~\cite{ng-etal-2020-ssmba}, GENIUS(-\textbf{ft})\footnote{\href{https://github.com/beyondguo/genius}{https://github.com/beyondguo/genius}} \cite{guo2022genius}, PromDA\footnote{\href{https://github.com/GaryYufei/PromDA}{https://github.com/GaryYufei/PromDA}}~\cite{wang-etal-2022-promda}, PromptMix\footnote{\href{https://github.com/servicenow/promptmix-emnlp-2023}{https://github.com/servicenow/promptmix-emnlp-2023}}~\cite{sahu2023promptmix}, ZeroGen\footnote{\href{https://github.com/jiacheng-ye/ZeroGen}{https://github.com/jiacheng-ye/ZeroGen}}~\cite{ye-etal-2022-zerogen}, GPT3Mix\footnote{\href{https://github.com/naver-ai/hypermix}{https://github.com/naver-ai/hypermix}}~\cite{yoo-etal-2021-gpt3mix-leveraging}, LwTR\footnote{\href{https://github.com/boschresearch/data-augmentation-coling2020}{https://github.com/boschresearch/data-augmentation-coling2020}}~\cite{dai-adel-2020-analysis}, DAGA\footnote{\href{https://github.com/ntunlp/daga}{https://github.com/ntunlp/daga}} \cite{ding-etal-2020-daga}\cite{ding-etal-2020-daga} and MELM\footnote{\href{https://github.com/randyzhouran/melm}{https://github.com/randyzhouran/melm}}~\cite{zhou2021melm}. All the baseline repositories are covered under the MIT License.

\subsection{Dataset Links}
We use the following datasets to evaluate: Huffpost\footnote{\href{https://www.kaggle.com/datasets/rmisra/news-category-dataset}{https://www.kaggle.com/datasets/rmisra/news-category-dataset}}~\cite{huffpost}, Yahoo\footnote{\href{https://huggingface.co/datasets/yahoo_answers_topics}{https://huggingface.co/datasets/yahoo\_answers\_topics}}~\cite{zhang2015character}, OTS\footnote{\href{https://huggingface.co/datasets/joelniklaus/lextreme/viewer/online_terms_of_service_unfairness_levels}{https://huggingface.co/datasets/joelniklaus/lextreme}}~\cite{drawzeski-etal-2021-corpus}, Massive\footnote{\href{https://huggingface.co/datasets/AmazonScience/massive/viewer/en-US}{https://huggingface.co/datasets/AmazonScience/massive}}~\cite{fitzgerald2022massive}, ATIS\footnote{\href{https://github.com/howl-anderson/ATIS_dataset}{https://github.com/howl-anderson/ATIS\_dataset}}~\cite{coucke2018snips}, ConLL-2003\footnote{\href{https://huggingface.co/datasets/conll2003}{https://huggingface.co/datasets/conll2003}}~\cite{tjong-kim-sang-de-meulder-2003-introduction},  OntoNotes-5.0\footnote{\href{https://catalog.ldc.upenn.edu/LDC2013T19}{https://catalog.ldc.upenn.edu/LDC2013T19}}~\cite{pradhan2013towards}, EBMNLP\footnote{\href{https://huggingface.co/datasets/bigbio/ebm_pico}{https://huggingface.co/datasets/bigbio/ebm\_pico}}~\cite{nye2018corpus} and BC2GM\footnote{\href{https://huggingface.co/datasets/bc2gm_corpus}{https://huggingface.co/datasets/bc2gm\_corpus}}~\cite{krallinger2015chemdner}, SQuAD\footnote{\href{https://rajpurkar.github.io/SQuAD-explorer/}{https://rajpurkar.github.io/SQuAD-explorer}}~\cite{rajpurkar2016squad} and NewsQA\footnote{\href{https://www.microsoft.com/en-us/research/project/newsqa-dataset/download/}{https://www.microsoft.com/en-us/research/project/newsqa-dataset/download/}}~\cite{trischler-etal-2017-newsqa}. All the datasets have been released under various licenses for research purposes.

\subsection{Potential Risks}
Diffusion models learn from vast amounts of textual data, including biased or prejudiced content present on the internet. As a result, there is a risk of bias amplification, where the models unintentionally perpetuate or reinforce existing biases. Also, diffusion models can generate highly coherent and contextually plausible text, raising concerns regarding the potential for generating misinformation or disinformation. 
\vspace{1mm}

\begin{table*}[]
\small
\centering
\setlength{\tabcolsep}{1.8pt}
{\renewcommand{\arraystretch}{1.5}%
\begin{tabular}{|p{0.05\linewidth}|p{0.25\linewidth}|p{0.35\linewidth}|p{0.35\linewidth}|}
\toprule
\multicolumn{1}{|c|}{\multirow{1}{*}{\textbf{Method}}} &  \multicolumn{1}{c|}{\textbf{Sentence}} &  \multicolumn{1}{c|}{\textbf{Instruction 1}}                     & \multicolumn{1}{c|}{\textbf{Instruction 2}}\\ \midrule 

Yahoo & Shops in most malls advertise for Christmas help up to the last minute. & Write a brief document with a single sentence or multiple sentences with the following constraints: 1. The document should have the following keywords: advertise or marketing, Shops, malls. 2. The document should be on the topic of Business \& Finance. 3. The document should have a length of 13-19 words. 4. Any sentence in the document should not include the abstract concept coaching. 5. Any sentence in the document should not include the abstract concept market volatility. 6. Any sentence in the document should not include the abstract concept market share. & Write a brief document with a single sentence or multiple sentences corresponding to the following abstract description: "Christmas help wanted ads in malls often run until the last minute." . Additionally, the document should have the following constraints: 1. The document should have the following keywords: business, industry, marketing, profits, but should not have the following keywords : develop. 2. The document should be on the topic of Business \& Finance. 3. The document should have a length of 413-619 words. 4. Any sentence in the document should not include the abstract concept coaching. 5. Any sentence in the document should not include the abstract concept market volatility. 6. Any sentence in the document should not include the abstract concept market share. \\

OTS & We are not obligated to publish any information or content on our Service and can remove it with or without notice. & Write a brief document with a single sentence or multiple sentences with the following constraints: 1. The document should have the following keywords: obligated,notice or prejudice,Service, but should not have the following keywords: responsible, liable. 2. The document's terms of service should be clearly unfair. 3. The document should have a part-of-speech sequence similar to: PRON AUX PART VERB PART VERB DET NOUN CCONJ NOUN ADP PRON PROPN CCONJ AUX VERB PRON ADP CCONJ ADP NOUN PUNCT. 4. The document should have a length of 21-31 words. 5. Any sentence in the document should not include the abstract concept litigation. 6. Any sentence in the document should not include the abstract concept account management. 7. Any sentence in the document should not include the abstract concept jurisdiction. & Write a brief document with a single sentence or multiple sentences corresponding to the following abstract description: "We reserve the right to remove content without notice.". Additionally, the document should have the following contraints: 1. The document should have the following keywords: reason,right,way or data, but should not have the following keywords: cause. 2. The document's terms of service should be clearly unfair. 3. The document should have a part-of-speech sequence similar to: PRON VERB DET NOUN PART VERB CCONJ VERB PROPN PROPN PUNCT CCONJ VERB DET NOUN PRON VERB ADP PROPN PUNCT ADP DET NOUN PUNCT. 4. The document should have a length of 21-31 words. 5. Any sentence in the document should not include the abstract concept litigation. 6. Any sentence in the document should not include the abstract concept account management. 7. Any sentence in the document should not include the abstract concept jurisdiction. \\

CoNLL-2003 & Israel approves Arafat's flight to West Bank. & Write a brief document with a single sentence or multiple sentences with the following constraints: 1. The document should have the following keywords: Israel,Arafat,West Bank,approves or confirms. 2. Israel is location, Arafat is person, West Bank is location. 3. The document should have a length of 5-13 words. & \\

BC2GM & Comparison with alkaline phosphatases and 5 - nucleotidase & Write a brief document with a single sentence or multiple sentences with the following constraints: 1. The document should have the following keywords: alkaline phosphatases,5 - nucleotidase,Comparison. 2. alkaline phosphatases is a Gene. 3. The document should have a part-of-speech sequence similar to: NOUN ADP ADJ NOUN CCONJ NUM PUNCT NOUN. 4. The document should have a length of 5-12 words. & \\\bottomrule           
\end{tabular}
}
\caption{\small Instruction prompts for various tasks.}
\label{tab:var}
\end{table*}

\begin{table*}[]
\small
\centering
\setlength{\tabcolsep}{1.8pt}
{\renewcommand{\arraystretch}{1.5}%
\begin{tabular}{|p{0.05\linewidth}|p{0.25\linewidth}|p{0.35\linewidth}|p{0.35\linewidth}|}
\toprule
\multicolumn{1}{|c|}{\multirow{1}{*}{\textbf{Method}}} &  \multicolumn{1}{c|}{\textbf{Sentence}} &  \multicolumn{1}{c|}{\textbf{Instruction 1}}                     & \multicolumn{1}{c|}{\textbf{Instruction 2}}\\ \midrule 

SQUAD & Beyoncé's first solo recording was a feature on Jay Z's "'03 Bonnie \& Clyde" that was released in October 2002, peaking at number four on the U.S. Billboard Hot 100 chart. Her first solo album Dangerously in Love was released on June 24, 2003, after Michelle Williams and Kelly Rowland had released their solo efforts. The album sold 317,000 copies in its first week, debuted atop the Billboard 200, and has since sold 11 million copies worldwide. The album's lead single, "Crazy in Love", featuring Jay Z, became Beyoncé's first number-one single as a solo artist in the US. The single "Baby Boy" also reached number one, and singles, "Me, Myself and I" and "Naughty Girl", both reached the top-five. The album earned Beyoncé a then record-tying five awards at the 46th Annual Grammy Awards; Best Contemporary R\&B Album, Best Female R\&B Vocal Performance for "Dangerously in Love 2", Best R\&B Song and Best Rap/Sung Collaboration for "Crazy in Love", and Best R\&B Performance by a Duo or Group with Vocals for "The Closer I Get to You" with Luther Vandross. & Write a brief document with multiple sentences corresponding to the following constraints: 1. The document should have the following keywords 11,Vocals,Hot,copies,lead,Baby,also,Vandross, You,Album,Best,earned,Rap/Sung,Grammy, Clyde,"Her first solo album Dangerously in Love was released on June 24, 2003, after Michelle Williams and Kelly Rowland had released their solo efforts". 2. The document should have a length of 113-340 words. &  \\

\bottomrule           
\end{tabular}
}
\caption{\small Instruction prompts for SQUAD dataset.}
\label{tab:squad}
\end{table*}

\end{document}